\documentclass[12pt]{article}
\usepackage[T1]{fontenc}
\usepackage{amsmath, amsfonts, bm}
\usepackage{graphicx}
\usepackage[hidelinks]{hyperref}
\usepackage{multirow}
\usepackage{fullpage}
\usepackage{subcaption}
\usepackage{setspace}

\title{Explainable Image Classification with Reduced Overconfidence for Tissue Characterisation}

\author{
Alfie Roddan$^{1}$,
Chi Xu$^{1}$,
Serine Ajlouni$^{2}$,
Irini Kakaletri$^{3}$,
Patra Charalampaki$^{2,4}$,\\
Stamatia Giannarou$^{1}$ \\
\\
$^1$The Hamlyn Centre for Robotic Surgery, Imperial College London, UK \\
$^2$Medical Faculty, University Witten Herdecke, Germany \\
$^3$Medical Faculty, Rheinische Friedrich Wilhelms University of Bonn, Germany \\
$^4$Department of Neurosurgery, Cologne Medical Center, Cologne, Germany \\
\\
\texttt{agr21@ic.ac.uk}
}

\date{}  

\begin{document}
\maketitle

\begin{abstract}
The deployment of Machine Learning models intraoperatively for tissue characterisation can assist decision making and guide safe tumour resections. For image classification models, pixel attribution methods are popular to infer explainability. However, overconfidence in deep learning model's predictions translates to overconfidence in pixel attribution. In this paper, we propose the first approach which incorporates risk estimation into a pixel attribution method for improved image classification explainability. The proposed method iteratively applies a classification model with a pixel attribution method to create a volume of PA maps. This volume is used for the first time, to generate a pixel-wise distribution of PA values. We introduce a method to generate an enhanced PA map by estimating the expectation values of the pixel-wise distributions. In addition, the coefficient of variation (CV) is used to estimate pixel-wise risk of this enhanced PA map. Hence, the proposed method not only provides an improved PA map but also produces an estimation of risk on the output PA values. Performance evaluation on probe-based Confocal Laser Endomicroscopy (pCLE) data and ImageNet verifies that our improved explainability method outperforms the state-of-the-art.
\end{abstract}

\section{Introduction}

When using a Machine Learning (ML) model during intraoperative tissue characterisation, it is vital that the surgeon trusts the output predictions of the model otherwise the model is rendered useless \cite{Diprose2020PhysicianCalculator}. For the surgeon to trust the output predictions of the model, the model must be able to explain itself \cite{Amann2020ExplainabilityPerspective}. One form of explainability in the image classification domain is pixel attribution (PA) mapping. PA maps aim to highlight the "most important" pixels to the classification. PA maps can be used to visually highlight whether a model is poorly extracting semantic features \cite{Zeiler2013VisualizingNetworks} and/or that the model is misinformed due to spurrious correlations within the data that it was trained on \cite{Hagos2022IdentifyingLearning}. To efficiently process image data, these methods mainly rely on Convolutional Neural Networks (CNNs) and achieve state-of-the-art (SOTA) performance. One of the first PA methods proposed for CNNs was class activation maps (CAM) \cite{ZhouLearningLocalization}. CAM uses one forward pass of the model to find the channels in the last convolutional layer that contributed most to the prediction. One of CAM's limitations is its reliance on global average pooling (GAP) \cite{Lin2013NetworkNetwork}  after the last CNN layer as it dramatically reduces the number of architectures that can use CAM. To improve on this, Grad-CAM \cite{Selvaraju2017Grad-CAM:Localization} generalises to all CNN architectures which are differentiable from the output logit layer to the chosen CNN layer. However, Grad-CAM often lacks sharpness in object localisation, as noted and improved on in Grad-CAM++ \cite{Chattopadhyay2017Grad-CAM++:Networks} and SmoothGrad-CAM++ \cite{Omeiza2019SmoothModels}. These extensions of Grad-CAM have good semantic feature localisation but they are unable to be deployed for use in surgery \cite{Byun2022Recipro-CAM:Map}. Both Score-CAM \cite{Wang2019Score-CAM:Networks} and Recipro-CAM \cite{Byun2022Recipro-CAM:Map} also generalise to all CNN architectures but are deployable. Score-CAM improves on object localisation within the visual PA map without losing the class specific capabilities of Grad-CAM by masking out regions of the image and measuring the change in the output score. This is similar to perturbation methods like RISE \cite{Petsiuk2018RISE:Models}, LIME \cite{Ribeiro2016WhyClassifier} and other perturbation techniques \cite{Zeiler2013VisualizingNetworks,Ancona2017TowardsNetworks}. On the other hand, Recipro-CAM focuses on the speed of PA map computation whilst maintaining comparable SOTA performance. By utilising the CNN's receptive field, Recipro-CAM generates a number of spatial masks and then measures the effect on the output score much like Score-CAM.

Despite being speedy, easy to deploy and able to localise semantic features, the above rely on the overconfident predictions of the underlying model. Deep learning (DL) models trained with empirical risk minimisation (ERM) are overconfident in prediction \cite{Gal2015DropoutLearning} and vulnerable to adverserial attacks \cite{Goodfellow2014ExplainingExamples}. Bayesian Neural Networks (BNNs) \cite{Neal1996BayesianNetworks} bring improved regularisation and output uncertainty estimates. Unfortunately, the non-linearity and number of variables within NNs make Bayesian inference a computationally intensive task. For this reason, variational methods \cite{Hinton1993KeepingWeights,GravesPracticalNetworks} are used to approximate Bayesian inference. More recently, the variational method Bayes by Backprop \cite{Blundell2015WeightNetworks} used Dropout \cite{Hinton2012ImprovingDetectors} to approximate Bayesian inference. Dropout is a regularisation technique which has also been noted to improve salient feature extraction. Although Bayes by Backprop is not overconfident, it often fails to scale to the complex architectures of SOTA models. To improve on this lack of generalisability, another variational method called Monte Carlo (MC) Dropout \cite{Gal2015DropoutLearning} proposes that a model trained with Dropout is equivalent to a probabilistic deep Gaussian process \cite{Damianou2012DeepProcesses,Gal2015DropoutAppendix}. With this assumption, an estimated output distribution is computed after a number of forward passes with Dropout have been applied. This output distribution is used in practice to indicate risk (uncalibrated variation) in the model's predictions. Using Dropout to perturb a model is a computationally cheap method of model averaging \cite{Hinton2012ImprovingDetectors}. It is worth noting though that this method's validity as a Bayesian Inference approximation was later questioned \cite{Folgoc2021IsBayesian}. However, this does not affect the use of this method for risk estimation. So far, model explainability and risk estimation have mostly been used separately to assess models' suitability for surgical applications.

In this paper, we propose the first approach which incorporates risk estimation into a PA method. 
A classification model is trained with Dropout and a PA method is used to generate a PA map. At test time, the classification model is employed with the Dropout enabled. In this work, we propose to repeat this process for a number of iterations creating a volume of PA maps. This volume is used for the first time, to generate a pixel-wise distribution of PA values from which we can infer risk. More specifically, we introduce a method to generate an enhanced PA map by estimating the expectation values of the pixel-wise distributions. In addition, the coefficient of variation (CV) is used to estimate pixel-wise risk of this enhanced PA map. This provides an improved explanation of the model's prediction by clearly presenting to the surgeon which salient areas to trust in the model's enhanced PA map. In this work, we focus on the explainability of the classification of brain tumours using probe-based Confocal Laser Endomicroscopy (pCLE) data but also demonstrate generalisation by evaluating on natural scenes. Performance evaluation on pCLE data shows that our improved explainability method outperforms the SOTA.

\section{Methodology}

\begin{figure}[t]
    \centering
    \includegraphics[width=\textwidth]{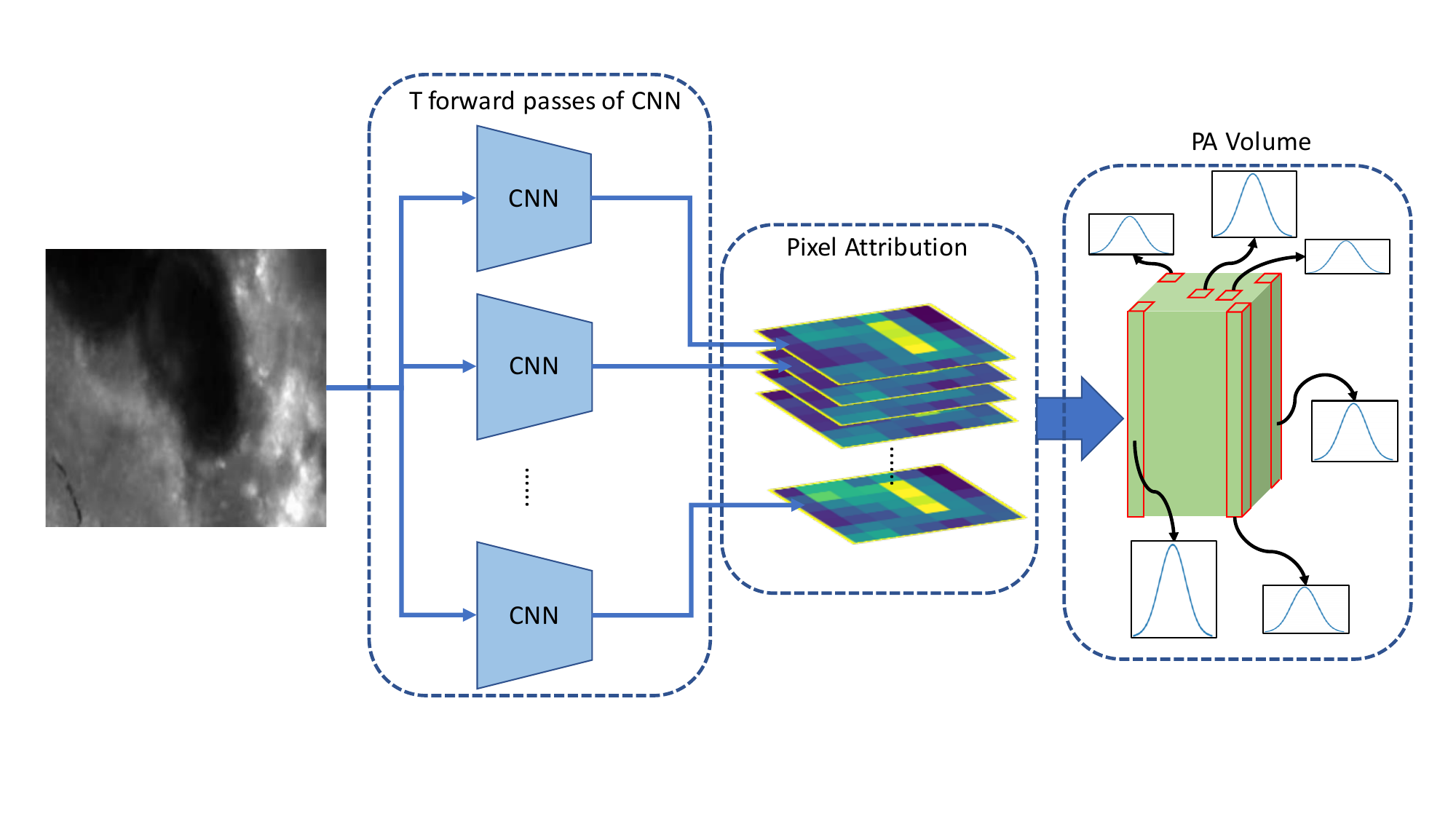}
    \caption{Outline of the proposed method. A PA volume is generated using T forward passes of a CNN model with Dropout applied.}
    \label{fig:method}
\end{figure}

The aim of the proposed method is to produce an improved PA map of a classification model, while providing risk estimation of the model's explainability. Further aiding the decision making during intraoperative tissue characterisation.

In our method, any CNN classification model trained with Dropout can be used. Let $\hat{\bm{Y}}$ be the output logits of the CNN model, where Dropout is enabled at test time, with input image $\bm{X} \in \mathbb{R}^{height \times width \times channels}$. 
Any PA method can be used to generate a PA map using the output logits $S = f_s(\hat{\bm{Y}}) \in \mathbb{R}^{height \times width}$ where $f_s(.)$ is the PA method. We propose to repeat the above process for $T$ iterations to create a volume of PA maps $\bm{S} = \{S_1,...,S_T\} \in \mathbb{R}^{height \times width \times T}$. We show this visually in Supplementary \ref{supp: D}.
A visual representation of how the volume is generated is show in Fig. \ref{fig:method}. 
The aim is to use this volume to generate a pixel-wise distribution of PA values from which we can infer risk. To achieve this, we compute the expectation and variance values of the volume along the third dimension as:

\begin{equation}
\begin{aligned}
    \mathbb{E}(\bm{S}_{i,j}) &\approx \frac{1}{T} \sum^{T}_{t=1} f_s(\hat{\bm{Y}}_{t})_{i,j} \\
    Var(\bm{S}_{i,j}) &\approx \frac{1}{T} \sum^T_{t=1} f_s(\hat{\bm{Y}}_{t})_{i,j}^T f_s(\hat{\bm{Y}}_{t})_{i,j} - \mathbb{E}(\bm{S}_{i,j})^T \mathbb{E}(\bm{S}_{i,j}),
\end{aligned}
\end{equation}
where, $i,j$ represent the pixel's row and column coordinates, respectively. The expectation $\mathbb{E}(\bm{S}_{i,j})$ of each pixel $(i,j)$ is used to generate an enhanced PA map of size $height \times width$. The intuition is that the above distribution of PA values can produce less noisy and overconfident estimation of a pixel's contribution to the final explainability map compared to a single estimate.

Advancing SOTA explainability methods, in our method we also estimate the risk of the enhanced PA map generated above. For the risk estimation, it is important to consider that different pixels in the PA map correspond to different semantic features which contribute differently to the output logits. This makes the pixe-wise distributions (and therefore expectation and variance values) to have different scales. For this purpose, the coefficient of variation (CV) is used to estimate pixel-wise risk, as it allows us to compare pixel-wise variances despite their different scales. This is mathematically defined as:

\begin{equation}
    S^{cv}_{i, j} = \frac{\sqrt{Var(\bm{S}_{i,j})}}{\mathbb{E}(\bm{S}_{i,j})} = \frac{std(\bm{S}_{i,j})}{\mathbb{E}(\bm{S}_{i,j})}.
    \label{eq: cv}
\end{equation}

\begin{figure}[t]
    \centering
    \includegraphics[width=\textwidth]{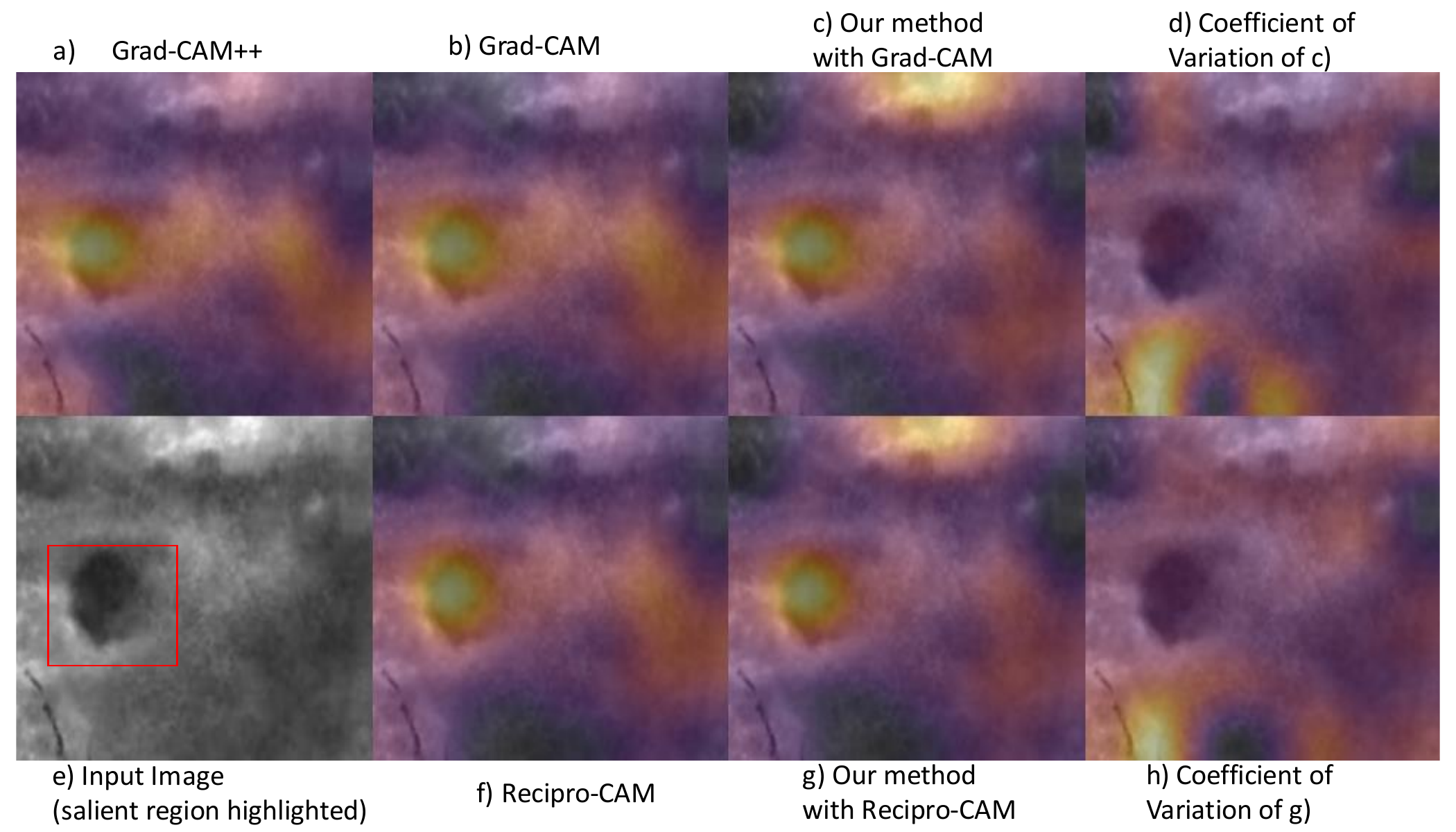}
    \caption{PA maps generated using ResNet18 on meningioma pCLE images. a) Grad-CAM++ PA map b) Grad-CAM PA map c) Grad-CAM PA map with our method applied d) Risk map (CV values) of c) e) original pCLE image with the salient region highlighted with red bounding box f) Recipro-CAM PA map g) Recipro-CAM PA map with our method applied h) Risk map (CV values) of g). Yellow represents the highest PA value and black the lowest.}
    \label{fig:viusal analayis_menin}
\end{figure}

Our proposed method improves ADDC and allows visualisation of both the explainability of the classification model (provided by the enhanced PA method) together with the pixel-wise risk of this map (provided by the CV map). For instance, salient areas on the PA map should not be trusted unless the CV values are low. An example of the enhanced PA and risk maps generated with the proposed method are shown in Figure \ref{fig:viusal analayis_menin}. This shows that the proposed method, not only improves explainability but also provides associated risk information which improves trustworthiness.

\section{Experiments and Analysis}

\subsubsection{Datasets} The developed explainability framework has been validated on an in vivo and ex vivo pCLE dataset of meningioma, glioblastoma and metastases of an invasive ductal carcinoma (IDC) collected at Anonymous Hospital. The Cellvizio© by Mauna Kea Technologies, Paris, France has been used in combination with the mini laser probe CystoFlex© UHD-R. The distinguishing characteristic of the meningioma is the psammoma body with concentric circles that show various degrees of calcification. Regarding glioblastomas, the pCLE images allow for the visualization of the characteristic hypercellularity, evidence of irregular nuclei with mitotic activities or multinuclear appearance with irregular cell shape. When examining metastases of an IDC, the tumor presents as egg-shaped cells with uniform evenly spaced nuclei. 
Our dataset includes 38 meningioma videos, 24 glioblastoma and 6 IDC. Each pCLE video represents one tumour type and corresponds to a different patient. The data has been curated to remove noisy images and similar frames. This resulted in a training dataset of 2500 frames per class (7500 frames in total) and a testing dataset of the same size. The dataset is split into a training and testing subset, with the division done on the patient level.To show generalisation to other domains, the proposed method was also evaluated on the ImageNet \cite{deng2009imagenet}database comprised of 1000 classes of natural scenes images.

\subsubsection{Implementation}
To implement the DL models we use the open-source framework PyTorch \cite{Paszke2019PyTorch:Library}, a NVIDIA Geforce RTX 3090 graphics card for parallel computation and a 12th Gen Intel(R) Core(TM) i9-12900K CPU (using 16 threads for latency experiments). To show our method generalises across domains we train and test on both pCLE data and ImageNet. For the pCLE data we train two lightweight models; ResNet-18 \cite{He2015DeepRecognition} and MobileNetV2 \cite{Sandler2018MobileNetV2:Bottlenecks}. ResNet-18 was trained with a learning rate of 0.001, whereas MobileNetV2 had a learning rate of 0.01 Both were trained using the Adam-W \cite{Loshchilov2017DecoupledRegularization} optimiser with a learning rate of 0.001 and weight decay of 0.01. For ImageNet we train a ResNet-50 with a learning rate of 0.1, using the Stochastic Gradient Descent optimizer \cite{article} and the Step learning rate scheduler with a step size 30 and gamma of 0.1. For the Resnet-50 model, we trained using distributed training on 3 X NVIDIA RTX A5000 graphics cards, evaluation was done on the NVIDIA Geforce RTX 3090 graphics card using the trained weights. All models were trained from scratch with a Dropout probability of 0.2 and a batch size of 256. At test time, we set $T=10$. PA methods were implemented with the help of TorchCAM \cite{Fernandez2020TorchCAM:Explorer}, ReciproCAM was implemented using the authors' source code.

\subsubsection{Evaluation Metrics} Evaluating a PA method is not a trivial task as a PA map may not need to be inline with what a human deems "reasonable" \cite{AdebayoSanityMaps}. Segmentation scores like intersection over union (IoU) may be used with caution to compare thresholded PA maps to ground truth maps with annotated salient regions. By doing so, we can measure how informed the model is about a particular class. To quantify how misinformed a model is, we can estimate at its average drop \cite{Chattopadhyay2017Grad-CAM++:Networks}:
\begin{equation}
    AverageDrop(f_s, \hat{\bm{Y}}, \bm{X}) = 100 \times \frac{max(0, \hat{\bm{Y}}(\bm{X}) - \hat{\bm{Y}}({\hat{\bm{X}}}))}{\hat{\bm{Y}}(\bm{X})},
    \label{eq: average drop}
\end{equation}
where, $\hat{\bm{X}} = \bm{X} \odot f_s(\hat{\bm{Y}}(\bm{X})$. The above equation measures the effect on the output score of the classification model if we only include the pixels which the PA method scored highly. A minimum average drop is desired. 

As average drop was found to not be sufficient on its own, the unified method ADCC \cite{Poppi2021RevisitingAnalysis} has been introduced which is the harmonic mean of average drop, coherency and complexity, defined as:
\begin{equation}
\label{eq: ADCC}
    \begin{aligned}
        ADCC(f_s(\hat{\bm{Y}})) = &(\frac{1}{Coherency(f_s(\hat{\bm{Y}}))}\\
        &+ \frac{1}{1-Complexity(f_s(\hat{\bm{Y}}))}\\
        &+ \frac{1}{1-AverageDrop(f_s, \hat{\bm{Y}}, \bm{X})})^{-1}.
    \end{aligned}
\end{equation}
Coherency is the Pearson Correlation Coefficient which ensures that the remaining pixels after dropping are still important, defined as:
\begin{equation}
    Coherency(f_s(\hat{\bm{Y}})) = 100 \times \frac{Cov(f_s(\hat{\bm{Y}}({\hat{\bm{X}}})), f_s(\hat{\bm{Y}}))}{\sigma(f_s(\hat{\bm{Y}}({\hat{\bm{X}}})) \sigma(f_s(\hat{\bm{Y}}))},
    \label{eq: coherency}
\end{equation}
where $Cov(., .)$ is the covariance. A higher coherency is better. Complexity is the L1 norm of the output PA map.
\begin{equation}
    Complexity(f_s(\hat{\bm{Y}}))) = 100 \times ||f_s(\hat{\bm{Y}}))||_1.
    \label{eq: comp}
\end{equation}
Complexity is used to measure how cluttered a PA map is. For a good PA map,  complexity should be a minimum. 
As it has been shown in the literature, the metrics in Eq. (\ref{eq: average drop}), (\ref{eq: coherency}) and (\ref{eq: comp}), can not be used individually to evaluate a PA method \cite{Poppi2021RevisitingAnalysis}. Whilst, ADCC combined with computation time gives us a reliable overall metric of how a PA method is performing.

\begin{table}[h]
    \centering
    \caption{ADCC vs Latency Study for ResNet-18 and MobilNetV2 on pCLE Dataset. Latency(ms) is the time to compute one PA map using a batch size of one.}\label{pCLE-adcc-latency}
    \begin{tabular}{l|cc|cc}
        \multicolumn{5}{c}{ResNet18}\\
        \hline
        & \multicolumn{2}{c|}{Original} & \multicolumn{2}{|c}{Proposed}\\
        \hline
        PA Method & ADCC $\uparrow$ & \multicolumn{1}{c|}{Latency $\downarrow$} & ADCC $\uparrow$ & Latency $\downarrow$\\
        \hline
        Grad-CAM & 76.6 & \textbf{75.6} & \textbf{77.7} & 9.7\\
        Grad-CAM++ & 76.2 & \textbf{5.5} & \textbf{78.2} & 10.5\\
        SmoothGradCAM++ & 74.8 & \textbf{70.7} & \textbf{75.6} & 103.9\\
        Score-CAM & \textbf{80.5} & \textbf{121.4} & \textbf{80.5} & 1267.2\\
        Recipro-CAM & 66.5 & \textbf{3.55} & \textbf{75.9} & 36.4\\
         \hline \hline

        \multicolumn{5}{c}{MobileNetV2}\\
        \hline
        & \multicolumn{2}{c|}{Original} & \multicolumn{2}{|c}{Proposed}\\
        \hline
        PA Method & ADCC $\uparrow$ & \multicolumn{1}{c|}{Latency $\downarrow$} & ADCC $\uparrow$ & Latency $\downarrow$\\
        \hline
        Grad-CAM & 29.3 & \textbf{8.8} &\textbf{48.0} & 12.5\\
        Grad-CAM++ & 37.8 & \textbf{8.7} & \textbf{59.5} & 13.4\\
        SmoothGradCAM++ & 24.5 & \textbf{71.3} & \textbf{37.1} & 88.8\\
        Score-CAM & \textbf{43.9} & \textbf{315.1} & \textbf{43.9} & 3154.3\\
        Recipro-CAM & 33.3 & \textbf{5.8} & \textbf{55.5} & 65.6\\
        \hline \hline
    \end{tabular}
\end{table}

\begin{table}[!ht]
    \centering
    \caption{ADCC vs Latency Study for ResNet-50 ImageNet Dataset. Latency(ms) is the time to compute one PA map using a batch size of one}\label{ImageNet-adcc-latency}
    \begin{tabular}{l|cc|cc}
        \multicolumn{5}{c}{ResNet50}\\
        \hline
        & \multicolumn{2}{c|}{Original} & \multicolumn{2}{|c}{Proposed}\\
        \hline
        PA Method & ADCC $\uparrow$ & \multicolumn{1}{c|}{Latency $\downarrow$} & ADCC $\uparrow$ & Latency $\downarrow$\\
        \hline
        Grad-CAM &  67.9 & \textbf{11.3} & \textbf{72.4} & 25.4\\
        Grad-CAM++ & 67.6 & \textbf{11.3} & \textbf{72.1} & 26.3\\
        SmoothGradCAM++ & 64.6 & \textbf{84.6} & \textbf{71.8} & 134.0\\
        Score-CAM & \textbf{74.3} & \textbf{134.0} & \textbf{74.3} & 14147.9\\
        Recipro-CAM & 63.0 & \textbf{7.5} & \textbf{70.9} & 86.6\\
        \hline \hline
    \end{tabular}
\end{table}

\subsection{T Study}

A parameter search was performed to find the optimal value of $T$. As show in Supplementary \ref{supp: C} there is a positive correlation between ADCC and the value of $T$. With increase of $T$, there is an implicit increase in latency. We found the optimal tradeoff of ADCC against latency to be $T=10$. The raw values of ADCC against T are in Supplementary \ref{supp: C}. 

\subsubsection{Performance Evaluation}

In a model's explanation we consider five metrics of performance; speed, usability, generalisability, trustworthiness and ability to localise semantic features. The proposed method has been compared to combinations of ResNet18 and MobileNetV2 with SOTA PA methods on both medical and natural scenes datasets. At test time, Dropout it not enabled for these standard methods (only for the proposed method). In Table \ref{pCLE-adcc-latency}, we show that our method outperforms all the compared CNN-PA method combinations on ADCC apart from Score-CAM. Much like our method, Score-CAM makes multiple forward passes on peturbed inputs, this makes it less susceptible to overconfidence. From Table \ref{ImageNet-adcc-latency} we show that this method generalises to the natural scenes domain. We believe that the better performance of our method is because of the random dropping of features taking place during Dropout at test time which helps to suppress noise in the estimated enhanced PA map. The combination of Recipro-CAM with our proposed method improves performance (increases ADCC) at the expense of increasing the computational complexity. We believe that this could be reduced using a batched implementation of Recipro-CAM. We attribute slow down in SmoothGradCAM++ when Dropout is applied during test time to the perturbations it adds on top of the PA method. Our validation study shows that Grad-CAM, Grad-CAM++ and Recipro-CAM are often leading in terms of speed as expected from the literature.

In Fig \ref{fig:viusal analayis_menin}, we can see our method includes more regions (top part) of the image and slightly sharper in localisation both of which would help the coherency and average drop metrics. Risk estimations from Eq. (\ref{eq: cv}) are also displayed and provide an added visualisation for a surgeon to evaluate both the model and the model's explanation. As it can be seen, areas of low CV match the areas of high PA values which shows the proposed explainability method is precise. During intraoperative surgery a surgeon can visualise an assitive DL model's explanation in order to assess or compare with what pixel's were found to be most relevant to the classfication. Whilst a PA method does not need to highlight a salient region, it does need to provide a fair and precise PA map for the surgeon to use. Our PA method removes overconfidence and also provides an added visualisation of relative precision, improving on SOTA PA methods.

\section{Conclusion} In this work we have introduced the first combination of risk in a PA methods. Using our proposed framework we not only improve on all the tested SOTA PA method's ADCC performances but also produce an estimation of risk on the output PA values. The proposed method can clearly present to the surgeon areas of the explainability map that are more trustworthy. From this work we hope to encourage trust between the surgeon and DL models by reducing overconfidence. For future work, we plan to deploy the proposed framework for use in surgery.

\subsubsection{Acknowledgements} 
This work was supported by the Engineering and Physical Sciences Research Council (EP/T51780X/1) and Intel R\&D UK. Dr Giannarou is supported by the Royal Society (URF$\setminus$R$\setminus$201014).

%
%
%
\bibliographystyle{plain}
\bibliography{my.bib}

\clearpage
\appendix
\section*{Supplementary Material}
\addcontentsline{toc}{section}{Supplementary Material}

\section{}
\label{supp: A}

\begin{table}[!ht]
    \centering
    \caption{Performance evaluation study on pCLE data based on the ADCC and time metrics. Coh is Coherence, Comp is Complexity, AD is average drop and they are reported for completeness. Time(s) is the average time to compute one PA map using a batch size of one.}\label{peformance-pcle}
    \begin{tabular}{l|lccccc}
        \hline 
        Architecture & PA method & \multicolumn{1}{|c}{Coh $\uparrow$} & Comp $\downarrow$ & AD $\downarrow$ & ADCC $\uparrow$ & Latency(ms) $\downarrow$\\ \hline
        \multirow{12}*{ResNet18}&\multicolumn{6}{c}{Original}\\ \cline{2-7}
        & Grad-CAM &\multicolumn{1}{|c}{90.1} & 32.7 & 10.1 & 76.6 & 5.6\\ 
        & Grad-CAM++ &\multicolumn{1}{|c}{90.6} & 33.1 & 10.6 & 76.2 & 5.5\\
        & SmoothGradCAM++ &\multicolumn{1}{|c}{88.3} & 27.6 & 14.3 & 74.8 & 70.7\\ 
        & Score-CAM &\multicolumn{1}{|c}{90.0} & 32.3 & 5.9 & 80.5 & 121.4 \\
        & Recipro-CAM &\multicolumn{1}{|c}{91.0} & 41.2 & 10.0 & 72.8 & 3.5 \\ \cline{2-7}
        &\multicolumn{6}{c}{Proposed method}\\ \cline{2-7}
        & Grad-CAM &\multicolumn{1}{|c}{92.5} & 34.2 & 11.9 & 77.7 & 9.7 \\ 
        & Grad-CAM++ &\multicolumn{1}{|c}{93.2} & 32.5 & 12.6 & 78.2 & 10.5 \\
        & SmoothGradCAM++ &\multicolumn{1}{|c}{92.2} & 30.6 & 17.2 & 75.6 & 103.9\\ 
        & Score-CAM &\multicolumn{1}{|c}{90.0}& 32.3 & 5.9 & 80.5 & 1267.2 \\
        & Recipro-CAM &\multicolumn{1}{|c}{92.1} & 37.8 & 11.8 & 75.9 & 36.4 \\ \hline \hline

        \multirow{12}*{MoblieNetV2}&\multicolumn{6}{c}{Original}\\ \cline{2-7}
        & Grad-CAM &\multicolumn{1}{|c}{89.5}  & 21.3 & 73.8 & 29.3 & 8.8\\ 
        & Grad-CAM++ &\multicolumn{1}{|c}{86.2} & 30.0 & 66.9 & 37.8 & 8.7 \\
        & SmoothGradCAM++ &\multicolumn{1}{|c}{77.7} & 18.1 & 76.2 & 24.5 & 71.3 \\ 
        & Score-CAM &\multicolumn{1}{|c}{62.5} & 33.9 & 56.3 & 43.9 & 315.1 \\
        & Recipro-CAM &\multicolumn{1}{|c}{85.8} & 32.3 & 67.1 & 35.8 & 5.8 \\ \cline{2-7}
        
        &\multicolumn{6}{c}{Proposed method}\\ \cline{2-7}
        & Grad-CAM &\multicolumn{1}{|c}{89.5} & 27.1 & 59.3 & 48.0 & 12.5\\ 
        & Grad-CAM++ &\multicolumn{1}{|c}{90.7} & 35.9 & 41.7 & 59.5 & 13.4\\
        & SmoothGradCAM++ &\multicolumn{1}{|c}{88.8} & 22.0 & 71.3 & 37.1 & 88.8 \\ 
        & Score-CAM &\multicolumn{1}{|c}{62.5}& 33.9 & 56.3 & 43.9 & 3154.3 \\
        & Recipro-CAM &\multicolumn{1}{|c}{90.2} & 33.8 & 48.6 & 55.5 & 65.6\\ \hline
    \end{tabular}
\end{table}

\clearpage
\section{}
\label{supp: B}

\begin{table}[!ht]
    \centering
    \caption{Performance evaluation study on ImageNet based on the ADCC and time metrics. Coh is Coherence, Comp is Complexity, AD is average drop and they are reported for completeness. Time(s) is the average time to compute one PA map using a batch size of one.}\label{peformance-imagenet}
    \begin{tabular}{l|lccccc}
        \hline 
        Architecture & PA method & \multicolumn{1}{|c}{Coh $\uparrow$} & Comp $\downarrow$ & AD $\downarrow$ & ADCC $\uparrow$ & Latency(ms) $\downarrow$\\ \hline
        \multirow{12}*{ResNet50}&\multicolumn{6}{c}{Original}\\ \cline{2-7}
        & Grad-CAM &\multicolumn{1}{|c}{98.1} & 34.0 & 30.0 & 67.9 & 11.3\\ 
        & Grad-CAM++ &\multicolumn{1}{|c}{98.3} & 34.8 & 30.0 & 67.6 & 11.28\\
        & SmoothGradCAM++ &\multicolumn{1}{|c}{97.3} & 34.8 & 34.8 & 64.6 & 84.6\\ 
        & Score-CAM &\multicolumn{1}{|c}{98.2} & 34.8 & 21.6 & 74.3 & 134.0 \\
        & Recipro-CAM &\multicolumn{1}{|c}{97.5} & 27.8 & 39.9 & 63.0 & 7.4 \\ \cline{2-7}
        &\multicolumn{6}{c}{Proposed method}\\ \cline{2-7}
        & Grad-CAM &\multicolumn{1}{|c}{97.7} & 35.2 & 22.3 & 72.4 & 25.4\\ 
        & Grad-CAM++ &\multicolumn{1}{|c}{97.7} & 35.8 & 22.6 & 72.1 & 26.3 \\
        & SmoothGradCAM++ &\multicolumn{1}{|c}{97.9} & 36.5 & 23.1 & 71.8 & 134.0\\ 
        & Score-CAM &\multicolumn{1}{|c}{98.2}& 34.8 & 21.6 & 74.3 & 14147.9 \\
        & Recipro-CAM &\multicolumn{1}{|c}{97.3} & 29.5 & 28.8 & 70.9 & 86.6 \\ \hline \hline
    \end{tabular}
\end{table}

\clearpage
\section{}
\label{supp: C}

\begin{figure}[!ht]
    \centering
    \includegraphics[width=\textwidth]{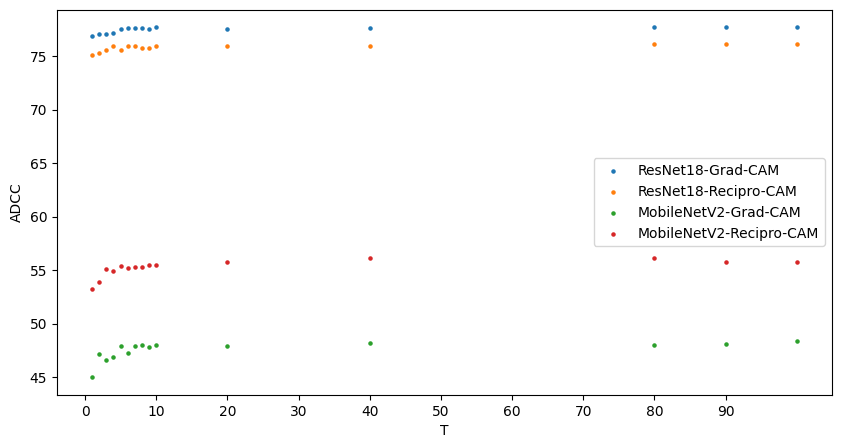}
    \caption{ADCC vs T parameter study for ResNet-18 and MobileNetV2 on pCLE dataset.}
    \label{fig:T study}
\end{figure}

\begin{table}[!ht]
    \centering
    \caption{T vs ADCC Study for Resnet-18 and MobileNetV2 on pCLE data.}\label{t-study}
    \begin{tabular}{c|c|c|c|c}
    \hline
     & \multicolumn{4}{c}{PA Method's ADCC} \\
     \hline
     & \multicolumn{2}{c}{ResNet18} & \multicolumn{2}{|c}{MobileNetV2} \\
     \hline
    T   & Grad-CAM & ReciproCAM & Grad-CAM                     & ReciproCAM \\
    \hline
    1   & 76.9     & 75.1       & 45                          & 53.2       \\
2   & 77.1     & 75.3       & 47.2                        & 53.9       \\
3   & 77.1     & 75.6       & 46.6                        & 55.1       \\
4   & 77.2     & 75.9       & 46.9                        & 54.9       \\
5   & 77.5     & 75.6       & 47.9                        & 55.4       \\
6   & 77.6     & 75.9       & 47.3                        & 55.2       \\
7   & 77.6     & 75.9       & 47.9                        & 55.3       \\
8   & 77.6     & 75.8       & 48                          & 55.3       \\
9   & 77.5     & 75.8       & 47.8                        & 55.5       \\
10  & 77.7     & 75.9       & 48                          & 55.5       \\
20  & 77.5     & 75.9       & 47.9                        & 55.8       \\
40  & 77.6     & 75.9       & 48.2                        & 56.1       \\
80  & 77.7     & 76.1       & 48                          & 56.1       \\
90  & 77.7     & 76.1       & 48.1                        & 55.8       \\
100 & 77.7     & 76.1  & 48.4 & 55.8 \\
    \hline \hline
    \end{tabular}
\end{table}

\clearpage
\section{}
\label{supp: D}

\begin{figure}[!ht]
    \centering
    \includegraphics[width=\textwidth]{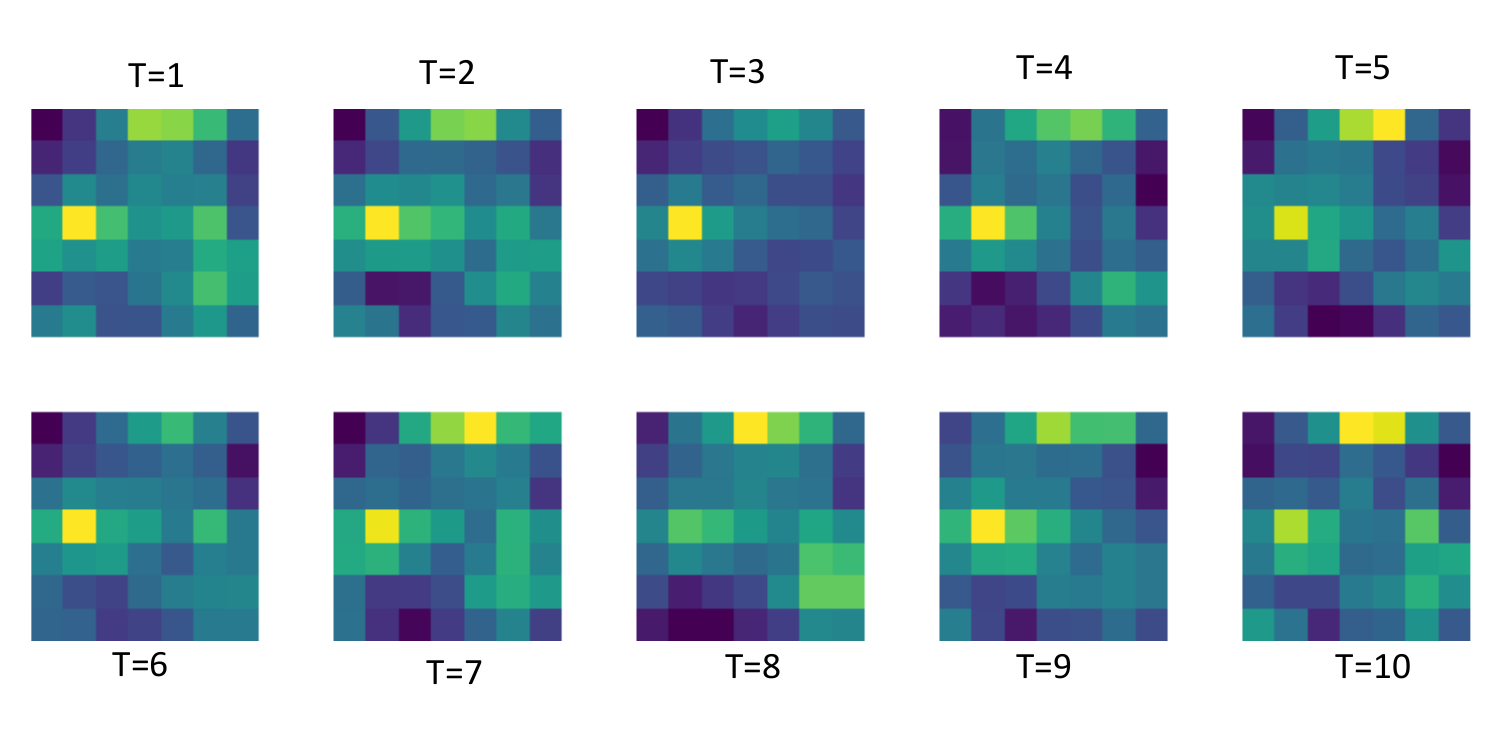}
    \caption{Each PA map of proposed method pre-interpolation to image size (for each iteration of T) from Layer 4 of ResNet18 for T=10 on pCLE data. Yellow represents the highest value and dark blue is lowest.}
    \label{fig:feature_maps}
\end{figure}

\clearpage
\section{}
\label{supp: E}

\begin{figure}[!ht]
    \centering
    \includegraphics[width=\textwidth]{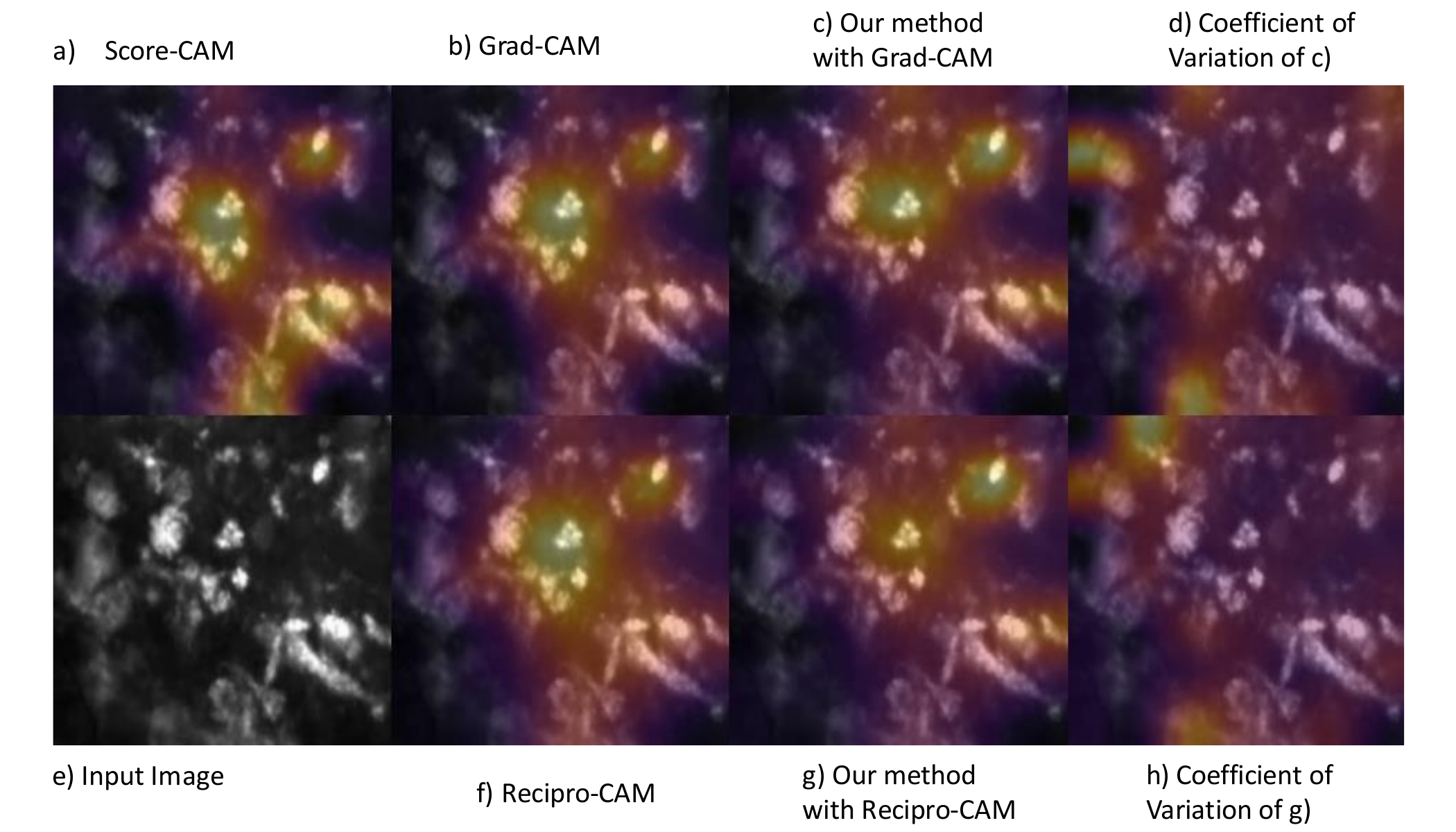}
    \caption{PA maps generated using ResNet18 on glioblastoma pCLE images. a) Score-CAM PA map b) Grad-CAM PA map c) Grad-CAM PA map with our method applied d) Risk map (CV values) of c) e) original pCLE image with the salient region highlighted with red bounding box f) Recipro-CAM PA map g) Recipro-CAM PA map with our method applied h) Risk map (CV values) of g). Yellow represents the highest PA value and black the lowest.}
    \label{fig:viusal analayis_glio}
\end{figure}

\begin{figure}[!ht]
    \centering
    \includegraphics[width=\textwidth]{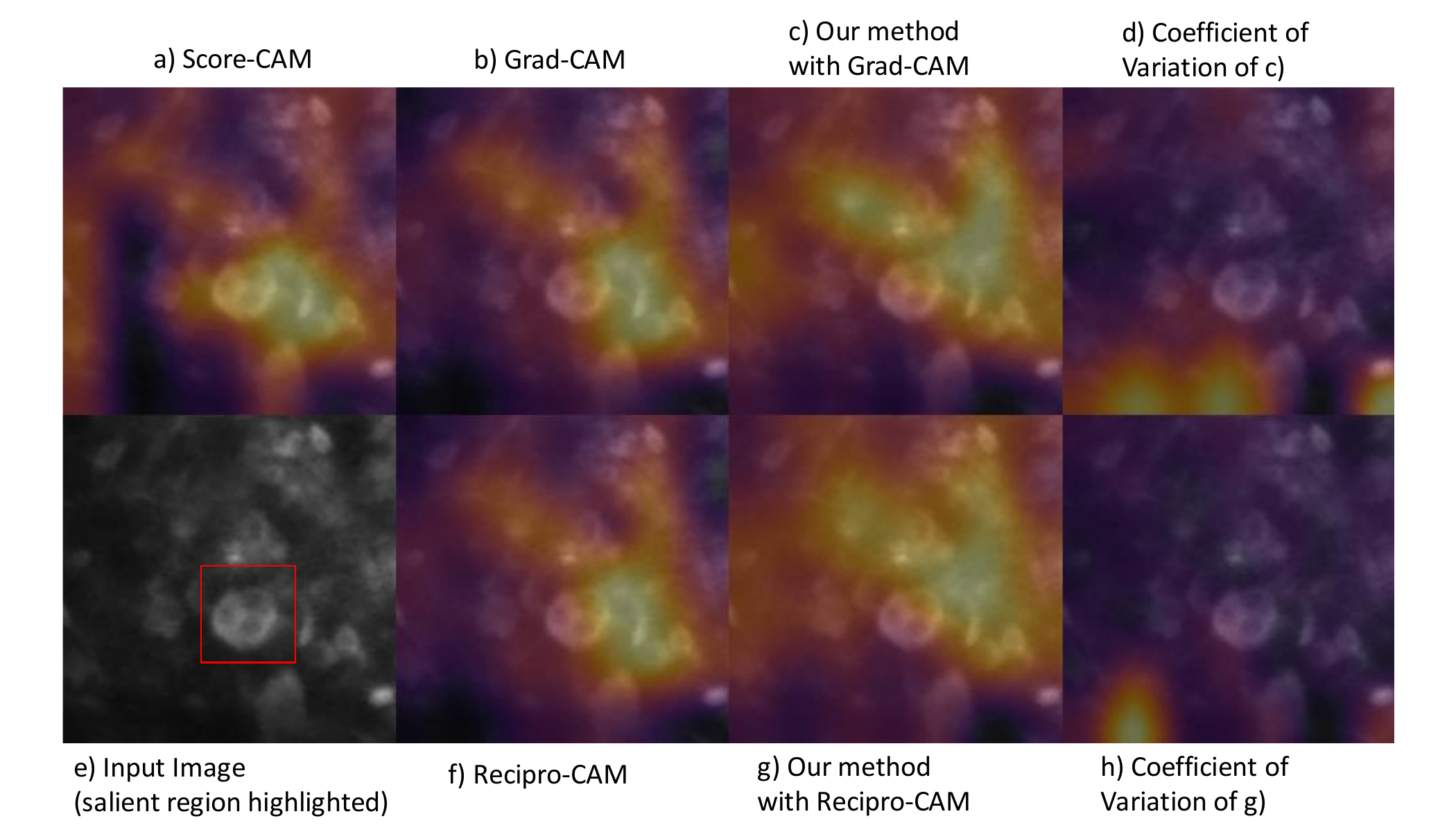}
    \caption{PA maps generated using ResNet18 on invasive ductal carcinoma (IDC) pCLE images. a) Score-CAM PA map b) Grad-CAM PA map c) Grad-CAM PA map with our method applied d) Risk map (CV values) of c) e) original pCLE image with the salient region highlighted with red bounding box f) Recipro-CAM PA map g) Recipro-CAM PA map with our method applied h) Risk map (CV values) of g). Yellow represents the highest PA value and black the lowest.}
    \label{fig:viusal analayis2}
\end{figure}

\begin{figure}[!ht]
    \centering
    \includegraphics[width=\textwidth]{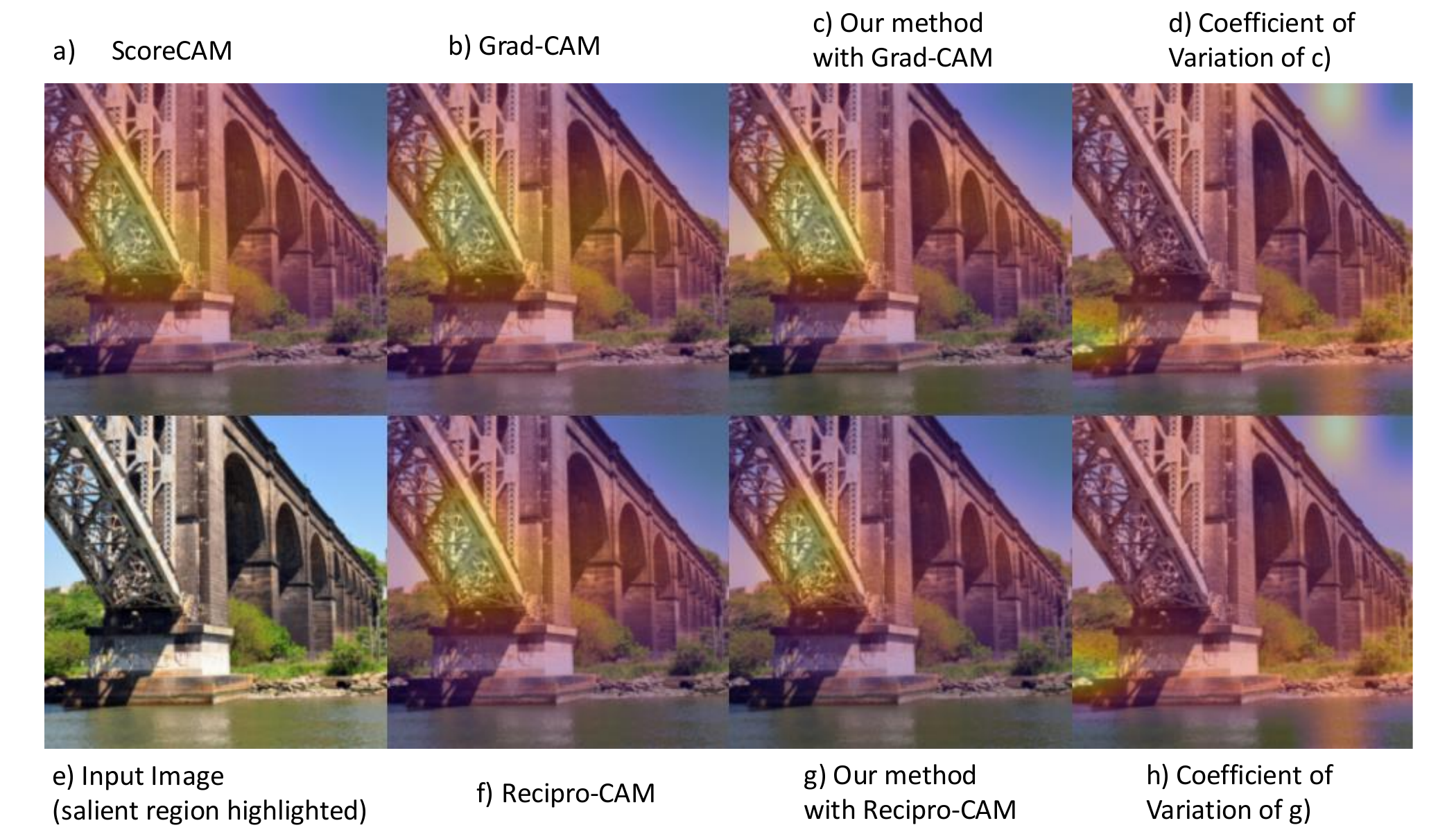}
    \caption{PA maps generated using ResNet50 on ImageNet. a) Score-CAM PA map b) Grad-CAM PA map c) Grad-CAM PA map with our method applied d) Risk map (CV values) of c) e) original image f) Recipro-CAM PA map g) Recipro-CAM PA map with our method applied h) Risk map (CV values) of g). Yellow represents the highest PA value and black the lowest.}
    \label{fig:viusal analayis_imagenet}
\end{figure}

\clearpage
\section{}
\label{supp: F}

\begin{figure}[!ht]
\centering
\begin{subfigure}{.3\textwidth}
    \centering
    \includegraphics[width=.95\linewidth]{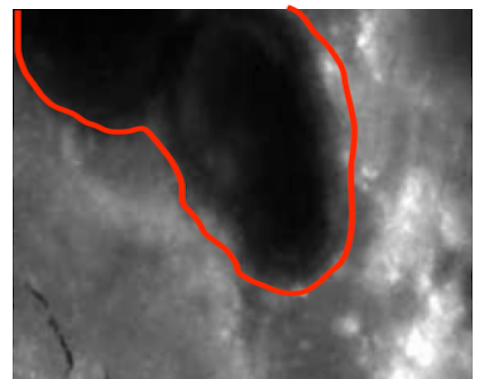}  
    \caption{Input}
    \label{exp1_input}
\end{subfigure}
\begin{subfigure}{.3\textwidth}
    \centering
    \includegraphics[width=.95\linewidth]{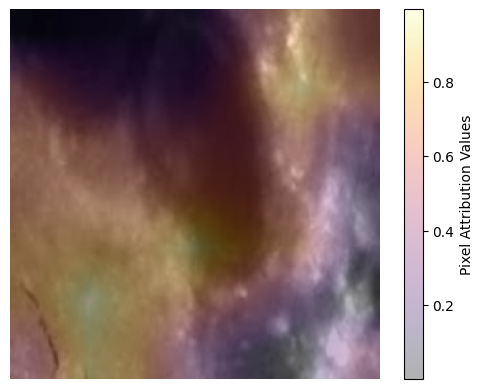}  
    \caption{GradCAM}
    \label{exp1_GradCAM}
\end{subfigure}
\begin{subfigure}{.3\textwidth}
    \centering
    \includegraphics[width=.95\linewidth]{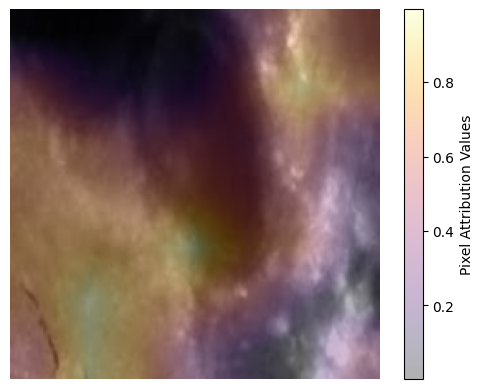}  
    \caption{GradCAMpp}
    \label{exp1_GradCAMpp}
\end{subfigure}
\begin{subfigure}{.3\textwidth}
    \centering
    \includegraphics[width=.95\linewidth]{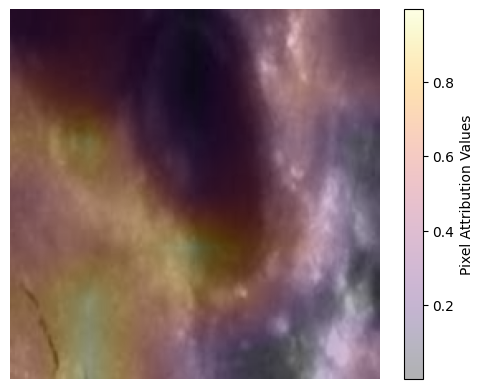}  
    \caption{SmoothGradCAMpp}
    \label{exp1_SmoothGradCAMpp}
\end{subfigure}
\begin{subfigure}{.3\textwidth}
    \centering
    \includegraphics[width=.95\linewidth]{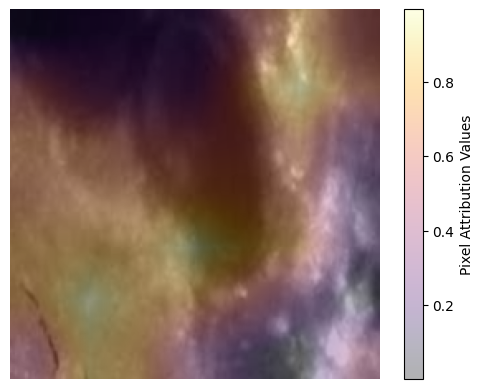}  
    \caption{ReciproCAM}
    \label{exp1_ReciproCAM}
\end{subfigure}
\begin{subfigure}{.3\textwidth}
    \centering
    \includegraphics[width=.95\linewidth]{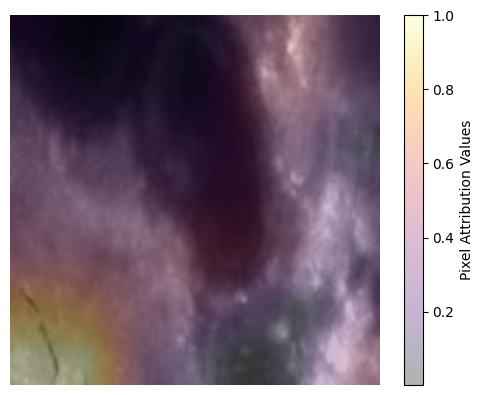}  
    \caption{ScoreCAM}
    \label{exp1_ScoreCAM}
\end{subfigure}
\caption{Multiple pixel attribution methods run on an un-robust model to shown spurious correlation of crack visible in Meningioma videos. Red annotation shows salient region psamoma body.}
\label{fig: spur_mng}
\end{figure}
\end{document}